\documentclass[letterpaper]{article}
\usepackage[preprint]{aaai2027}
\usepackage[hyphens]{url}
\usepackage{graphicx}
\usepackage{natbib}
\usepackage{caption}
\usepackage{booktabs}
\usepackage{amsmath}
\usepackage{amssymb}

\title{RATL: Learning from Retrieved Residuals for Robust Multivariate Time-Series Forecasting}
\author{
Yuchen He\textsuperscript{\rm 1},
Yueyang Cang\textsuperscript{\rm 1},
Zhiyuan Ning\textsuperscript{\rm 1},\\
Ningyu Wang\textsuperscript{\rm 2}\corresponding,
Li Shi\textsuperscript{\rm 1}\corresponding
}
\affiliations{
\textsuperscript{\rm 1}Department of Automation, Tsinghua University\\
\textsuperscript{\rm 2}State Key Laboratory of Hydroscience and Engineering, Tsinghua University\\
heyuchen25@mails.tsinghua.edu.cn,
cangyy23@mails.tsinghua.edu.cn,
ningzy25@mails.tsinghua.edu.cn\\
wang-ningyu@mail.tsinghua.edu.cn,
shilits@tsinghua.edu.cn
}

\begin{document}
\maketitle

\begin{abstract}
Retrieval-augmented generation (RAG) complements parametric models with retrieved external evidence. The same idea is attractive for continuous-output regression, but directly reusing retrieved target values is often not robust when samples differ in output level, numerical scale, or local dynamics. Moreover, conventional forecasting pipelines generally use residuals for model optimization and error diagnosis, but do not retain individual historical residual examples as memory that can be accessed at inference time.
For multivariate time-series forecasting, we propose RATL, a plug-in residual-retrieval and feedback-correction method. RATL freezes a base forecaster to construct retrieval keys and turns its historical forecast residuals into a train-only memory specific to that base model. At inference time, RATL retrieves residual trajectories from similar historical contexts subject to causal availability constraints, then uses a set-aware router operating over forecast blocks and variables to select and combine these trajectories. Experiments show that historical residuals matched to the current context contain reusable forecasting information and that RATL improves frozen base forecasters in most experimental settings. Ablations further show that learned routing strengthens raw residual feedback, while validation-based correction-strength selection limits residual over-injection.
On real-world benchmarks, we use iTransformer as the primary frozen base forecaster, compare against multiple strong forecasting baselines, and test transferability across backbones. The results show that RATL can further improve base-forecaster performance in most settings.
Overall, RATL shifts the retrieved object from historical target values to base-model-specific historical forecast errors, providing a plug-in, residual-memory-based paradigm for learned feedback correction in continuous-output forecasting.
\end{abstract}

\section{Introduction}

Long-horizon multivariate forecasting maps a local context window to a future trajectory. Modern linear, MLP, and Transformer forecasters have improved this mapping substantially \cite{zeng2023dlinear,nie2023patchtst,liu2024itransformer}, yet their prediction is still conditioned on a fixed lookback and on historical patterns compressed into learned parameters. When a relevant pattern occurred far outside the lookback, a parametric forecaster cannot explicitly inspect that occurrence at inference time.

Retrieval-augmented forecasting addresses this limitation by searching a historical datastore and incorporating the futures associated with similar contexts \cite{zhang2025knnmts,han2025raft,du2026pfrp}. However, raw future reuse creates two coupled risks. First, similar contexts need not share the same future level or scale. Second, an imperfect retrieved signal may be injected even when the base prediction is already accurate, causing \emph{negative transfer}. Both issues become pronounced in multivariate, long-horizon settings, where the correction must vary across variables and forecast blocks.

\begin{figure*}[t]
    \centering
    \includegraphics[width=\textwidth]{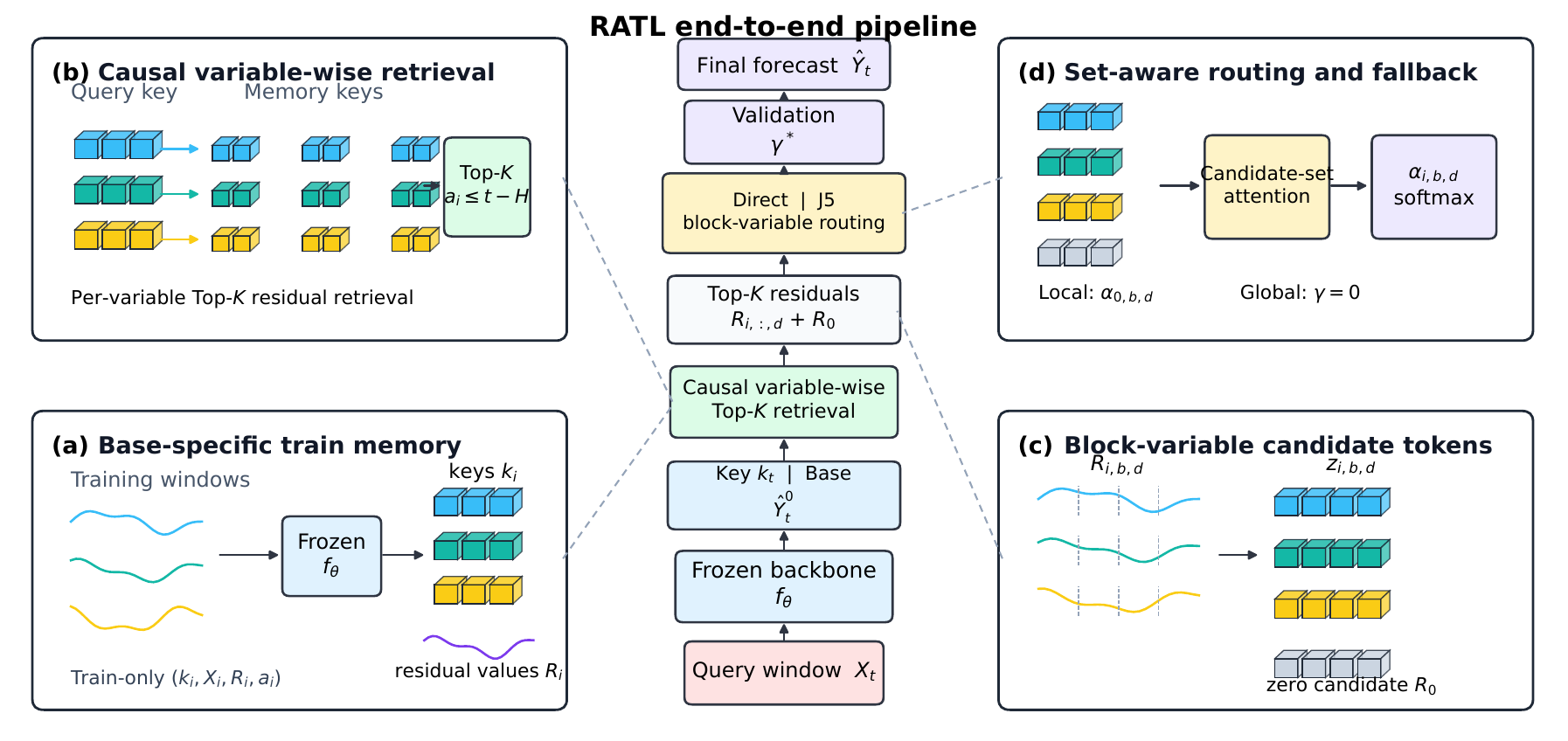}
    \caption{Overview of RATL. The center shows the end-to-end pipeline from an input window to the final corrected forecast; (a) a frozen base forecaster constructs a train-only key--residual memory; (b) Top-$K$ residuals are retrieved independently for each variable under the temporal-availability constraint; (c) candidate trajectories are split into block--variable tokens and augmented with a zero-residual candidate; (d) J5 predicts candidate weights through set attention. Direct similarity weighting is the non-parametric baseline and uses the raw correction with $\gamma=1$; for RATL, the correction-strength hyperparameter $\gamma$ is selected on validation before one-shot test evaluation.}
    \label{fig:overview}
\end{figure*}

We investigate a different retrieval target: the \emph{forecast error of a fixed base model}. For each training window, we store the base-model residual rather than treating the observed future as a stand-alone prediction, yielding an output-space correction. RATL retrieves only historical residuals whose target windows are temporally available from the training memory. In its main configuration, RATL uses the frozen base forecaster's per-variable hidden representations as retrieval keys. J5, a set-aware learned residual router, constructs block--variable candidate tokens from the query, candidate residual blocks, a Direct residual reference, and positional features. Set attention assigns weights to retrieved residual candidates and an explicit zero-residual candidate, while a global correction-strength hyperparameter $\gamma\in[0,1]$ is selected on validation. As a control, the non-parametric \emph{Direct} baseline uses only retrieval similarity as weights for a raw weighted correction over the same residual candidates, with $\gamma=1$ and no separate correction-strength tuning. Figure~\ref{fig:overview} presents the overall design: the central pipeline consists of frozen-base forecasting, causal residual retrieval, block--variable routing, and correction-strength selection, while four side panels expand its key operations.

Experiments show that RATL can turn context-matched historical errors into useful corrections across different forecasting settings and improves overall forecasting performance over the Direct baseline.

Our contributions are:
\begin{itemize}
    \item We formulate retrieval-augmented forecasting as \emph{base-specific residual retrieval}, with a train-only memory and an explicit temporal admissibility rule that prevents target leakage.
    \item We develop RATL around J5, a set-aware, block-variable residual router with oracle imitation and a no-correction candidate; similarity-weighted Direct serves as the non-parametric retrieval baseline.
    \item We conduct a 156-run study covering 13 datasets and 52 settings, together with audited transfer studies on frozen DLinear, PatchTST, TimesNet, and TimeMixer checkpoints. RATL improves mean MSE by 9.57\% in the main study. The transfer results show that its gains vary substantially with the base-forecaster architecture, dataset, and retrieval-key choice.
\end{itemize}

\section{Related Work}

\paragraph{Multivariate forecasting.}
Informer reduces attention cost for long sequences \cite{zhou2021informer}; Autoformer and FEDformer introduce decomposition and frequency-aware modeling \cite{wu2021autoformer,zhou2022fedformer}; TimesNet models temporal variation in two dimensions \cite{wu2023timesnet}; and TimeMixer decomposes multiscale patterns through mixing blocks \cite{wang2024timemixer}. More recent strong baselines include the decomposition-based DLinear \cite{zeng2023dlinear}, channel-independent PatchTST \cite{nie2023patchtst}, and iTransformer, which represents variables as tokens \cite{liu2024itransformer}. RATL does not replace these architectures: it is trained after a base forecaster and operates in prediction space. Our main experiments use iTransformer; separate frozen audits attach the same correction interface to DLinear, PatchTST, TimesNet, and TimeMixer checkpoints.

\paragraph{Retrieval-augmented forecasting.}
Retrieval augmentation has been effective in language modeling by exposing non-parametric memory at inference time \cite{lewis2020rag,khandelwal2020knnlm}. In time series, kNN-MTS retrieves future segments using learned multivariate representations \cite{zhang2025knnmts}; RAFT retrieves matching historical patches at multiple periods \cite{han2025raft}; and PFRP combines global retrieved predictions with a local univariate model using confidence and output gates \cite{du2026pfrp}. Foundation-model variants concatenate or prompt with retrieved contexts \cite{tire2024raf}. RATL differs in the memory value and in the object being combined: it stores the \emph{residual of a specific frozen forecaster}, then learns to select and compose an additive correction for multivariate blocks. This makes the retrieved value a model-failure template rather than a stand-alone future.

Table~\ref{tab:retrieval-related} summarizes the differences between RATL and representative retrieval-augmented forecasting methods along the retrieval-object and fusion dimensions.
\begin{table}[t]
\centering
\small
\setlength{\tabcolsep}{3pt}
\begin{tabular}{lll}
\toprule
Method & Retrieved value & Integration \\
\midrule
kNN-MTS & future segment & similarity aggregation \\
RAFT & future patches & input/feature augmentation \\
PFRP & global prediction & confidence/output gates \\
RATL & frozen-base residual & raw Direct / J5 + val.\ strength \\
\bottomrule
\end{tabular}
\caption{Retrieval target and integration differ across related forecasters.}
\label{tab:retrieval-related}
\end{table}

\paragraph{Residual modeling and negative transfer.}
Residual structure has long been exploited by hybrid statistical--neural models and boosting \cite{zhang2003hybrid,friedman2001greedy}. ResMem fits a nearest-neighbor regressor to a base model's training residuals \cite{yang2023resmem}, while similarity-based macroeconomic forecasting corrects ARIMA predictions with errors from related historical periods \cite{guerron2023macroeconomic}. MSCT-RCM constructs a KNN residual-sequence library for ultra-short photovoltaic forecasting \cite{ye2026multiscale}, whereas $\delta$-Adapter learns a bounded post-processing correction for frozen forecasters without retrieving historical residuals \cite{liang2026forecast}. RATL extends the residual-reuse principle to multivariate long-horizon trajectories, causal train-only retrieval, and block--variable routing, and uses a zero-residual candidate and a correction-strength hyperparameter to suppress noise in retrieved historical residuals and reduce forecast error.

\section{Method}

\subsection{Problem Setup and Frozen Base}

Let $\mathbf{X}_t\in\mathbb{R}^{L\times D}$ be a lookback window ending at forecast origin $t$ and $\mathbf{Y}_t\in\mathbb{R}^{H\times D}$ its future. A base forecaster $f_\theta$ produces
\begin{equation}
    \widehat{\mathbf{Y}}^{\,0}_t=f_\theta(\mathbf{X}_t),
    \qquad
    \mathbf{R}_t=\mathbf{Y}_t-\widehat{\mathbf{Y}}^{\,0}_t .
\end{equation}
We first train $f_\theta$ conventionally and then freeze it. RATL learns only to estimate a correction $\widehat{\mathbf{R}}_t$:
\begin{equation}
    \widehat{\mathbf{Y}}_t
    =\widehat{\mathbf{Y}}^{\,0}_t+\gamma\widehat{\mathbf{R}}_t ,
    \qquad 0\leq\gamma\leq1.
    \label{eq:corrected-forecast}
\end{equation}
This separation lets the same residual-retrieval interface wrap different forecasters and makes every memory value interpretable as a historical error of the deployed base.

\subsection{Train-Only Residual Memory}

For each training window $i$, we store
\begin{equation}
    \mathcal{M}_i=(\mathbf{k}_i,\mathbf{X}_i,\mathbf{R}_i,a_i),
    \qquad \mathbf{k}_i=\phi_\theta(\mathbf{X}_i),
\end{equation}
where $\phi_\theta$ is a frozen base representation and $a_i=t_i+H$ is the time at which the full target used to compute $\mathbf{R}_i$ becomes available. In the main configuration, $\phi_\theta$ is the iTransformer encoder's variable-token representation. The frozen transfer protocols separately use a backbone-agnostic input-statistics key for DLinear/PatchTST and preregistered native hidden keys for PatchTST, TimesNet, and TimeMixer; no key is selected using test results. Search uses squared Euclidean similarity independently for variable tokens.

\paragraph{Why retrieve residuals?}
A retrieved future can be decomposed as $\mathbf{Y}_i=f_\theta(\mathbf{X}_i)+\mathbf{R}_i$. Reusing $\mathbf{Y}_i$ asks retrieval to transfer both the neighbor's forecastable level and its model error. RATL retains only $\mathbf{R}_i$: the current base remains responsible for level, trend, and cross-variable structure, while memory estimates the systematic component that the same model missed in a related context. The memory is therefore base-specific. Replacing $f_\theta$ requires rebuilding residual values, but not redesigning the retrieval/correction interface.

\subsection{Per-Variable Top-$K$ Search with Retrieval Keys}

Given a current input window $\mathbf{X}_t$, we compute the query key $\mathbf{k}_t=\phi_\theta(\mathbf{X}_t)$ using the same frozen mapping used to build the memory. For a per-variable key, $\mathbf{k}_{t,d}\in\mathbb{R}^{P}$ is the query representation of variable $d$ over the full lookback window. Retrieval is performed independently for each variable, rather than separately for future forecast blocks.

Retrieval first applies a temporal-availability constraint. At forecast origin $t$, candidate $i$ may participate in the search only if
\begin{equation}
    a_i \leq t-H .
    \label{eq:temporal-mask}
\end{equation}
Because $a_i=t_i+H$ is the time at which the candidate target becomes fully available, this constraint inserts an additional gap of length $H$ after the candidate target, preventing retrieval of memory entries that are too close to the current forecast or whose future information is not yet available.

In the frozen main protocol, keys receive no additional normalization, and retrieval uses the per-variable mean squared Euclidean distance
\begin{equation}
    d_{t,i,d}
    =\frac{1}{P}\left\|\mathbf{k}_{t,d}-\mathbf{k}_{i,d}\right\|_2^2,
    \qquad
    s_{t,i,d}=-d_{t,i,d}.
\end{equation}
A smaller distance gives a larger retrieval score $s_{t,i,d}$. For each variable, we select only the $K$ highest-scoring entries among the temporally admissible candidates:
\begin{equation}
    \mathcal{N}_{t,d}
    =\operatorname{TopK}_{i:\,a_i\leq t-H}
      \bigl(s_{t,i,d}\bigr).
\end{equation}
Retrieval depends only on the current query key, historical memory keys, and the temporal-availability constraint; it does not use the current ground-truth future or current true residual. After the search, we take the full residual trajectories $\{\mathbf{R}_{i,:,d}\}_{i\in\mathcal{N}_{t,d}}$ associated with these $K$ historical windows. Direct combines these trajectories directly according to their retrieval scores, whereas J5 learns candidate weights within the same candidate set for each time block and variable. Thus, Top-$K$ search operates on representations of the full input window, and forecast blocking occurs only after retrieval.

\subsection{Direct Similarity-Weighted Baseline}

The Direct corrector converts retrieval scores $s_{t,i,d}$ into weights
\begin{equation}
    \pi_{t,i,d} =
    \frac{\exp(s_{t,i,d}/\tau_s)}
         {\sum_{j\in\mathcal{N}_{t,d}}\exp(s_{t,j,d}/\tau_s)}
\end{equation}
and averages the corresponding residual trajectories:
\begin{equation}
    \widehat{\mathbf{R}}^{\,\mathrm{dir}}_{t,:,d}
    =\sum_{i\in\mathcal{N}_{t,d}}\pi_{t,i,d}\mathbf{R}_{i,:,d}.
    \label{eq:direct}
\end{equation}
Direct is parameter-free after the base is trained and serves as the non-parametric control for RATL. It tests whether learned routing improves over simply averaging the same retrieved candidates and residual values by context similarity. Direct is evaluated as this raw correction with $\gamma=1$ and is not separately tuned for correction strength.

\subsection{Block-Residual Candidates and the Soft-Oracle Teacher}

The retrieval module returns $K$ historical residual trajectories for the current query. Using one candidate weight for an entire trajectory is too coarse because the same historical residual may be useful early in the forecast but fail later; selecting at every time point is more susceptible to local noise and would create $H\times D$ groups of fine-grained decisions. To balance flexibility and stability, we partition the forecast horizon into consecutive temporal blocks of length $B_h=8$. Let $\mathcal{B}_b$ denote the forecast positions in block $b$, and let the number of blocks be $G=\lceil H/B_h\rceil$. J5 assigns candidate-residual weights separately for every temporal block $b$ and variable $d$, allowing one candidate to correct only part of the forecast interval or only some variables.

During training, the target $\mathbf{Y}_t$ is known, so the true residual of the frozen base forecaster on the current query can be computed as
\[
    \mathbf{R}_t=\mathbf{Y}_t-\widehat{\mathbf{Y}}^{\,0}_t.
\]
For each retrieved candidate $i\in\{1,\ldots,K\}$, we compare its historical residual $\mathbf{R}_i$ with the current true residual $\mathbf{R}_t$ at block--variable position $(b,d)$. We also define candidate $i=0$ as the zero residual, $\mathbf{R}_{0,b,d}=\mathbf{0}$, representing no correction. The candidate error is
\begin{equation}
    e_{t,i,b,d}
    =\frac{1}{|\mathcal{B}_b|}\sum_{h\in\mathcal{B}_b}
      \left(R_{i,h,d}-R_{t,h,d}\right)^2.
\end{equation}
A smaller error means that candidate $i$ more closely matches the correction that the base forecaster actually needs for that time block and variable. We therefore construct the soft Oracle teacher distribution
\begin{equation}
    q^*_{t,i,b,d}
    =\operatorname{softmax}_i(-e_{t,i,b,d}/\tau_o),
\end{equation}
where $\tau_o$ controls the smoothness of the teacher distribution. Compared with a hard label that selects only the minimum-error candidate, the soft distribution can retain probability mass on several similarly effective candidates and reduce supervision instability caused by small fluctuations in candidate error. This Oracle is used only to construct the supervision target during training; the true future is unavailable at validation and test time, so $q^*$ is neither computed nor used then.

\begin{figure*}[t]
    \centering
    \includegraphics[width=\textwidth]{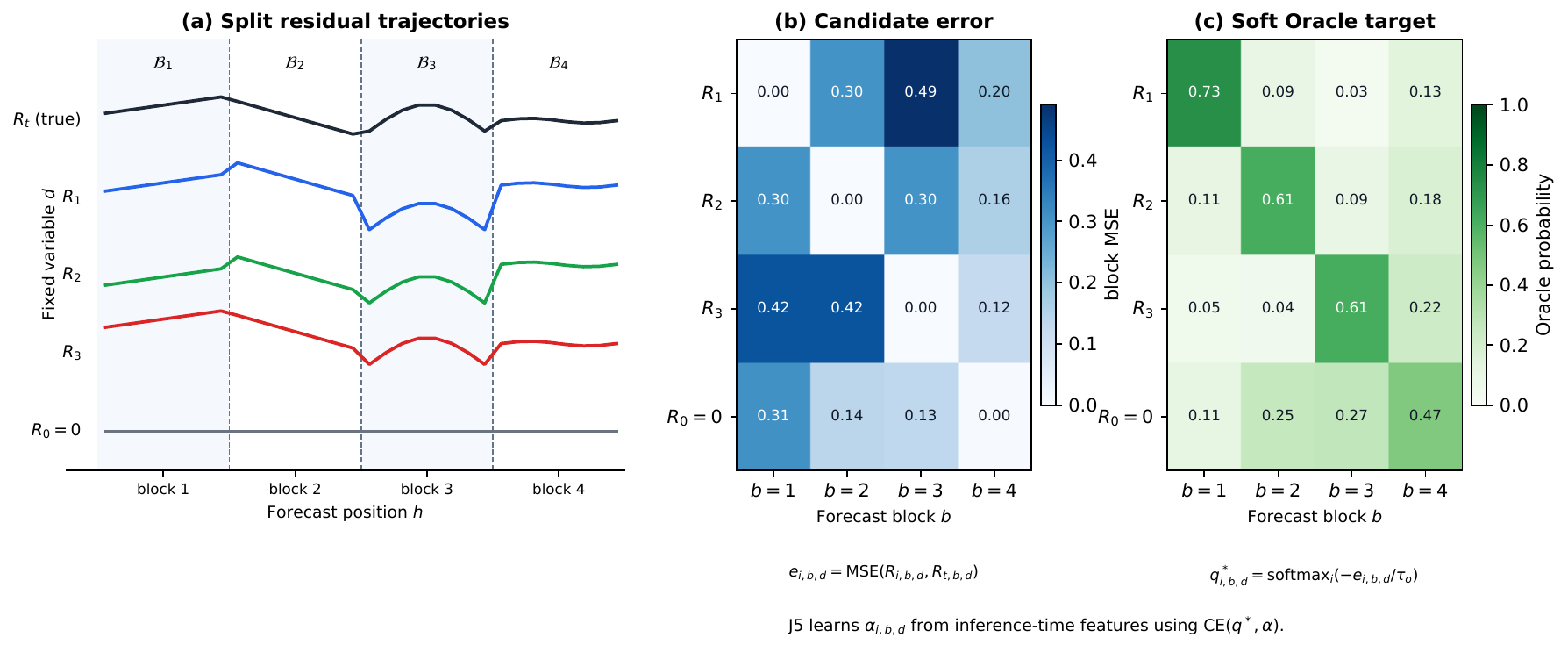}
    \caption{Construction of block-residual candidates and the soft Oracle supervision target for a fixed variable $d$. Full residual trajectories are first divided into consecutive temporal blocks along the forecast horizon. In each block, every historical candidate and the zero-residual candidate $R_0=0$ are compared with the current true residual to obtain the block--variable error $e_{i,b,d}$. A temperature-scaled softmax over negative errors then gives the teacher distribution $q^*_{i,b,d}$. Curves and values in the figure illustrate the procedure and are not experimental results.}
    \label{fig:block-oracle}
\end{figure*}

Figure~\ref{fig:block-oracle} illustrates the ``trajectory blocking--candidate comparison--soft teacher distribution'' construction for a single variable.

\subsection{RATL Set-Aware Residual Router}

Direct assumes that context similarity directly represents residual utility. J5 instead predicts the block--variable Oracle distribution above, without observing the current true residual, from information available at inference time. For candidate $i$, temporal block $b$, and variable $d$, the router constructs a candidate token from the current query, candidate residual block, Direct residual reference, and block/variable positional features. The general router also supports query--neighbor window relations and similarity features; the frozen main variant disables similarity input/prior and masks the neighbor-window, window-difference, residual-mean, and Direct-residual-energy channels specified by its feature-ablation mask. Its active token can be written
\begin{equation}
    \mathbf{z}_{t,i,b,d}
    =g\!\left(\mathbf{X}_t,
      \mathbf{R}_{i,b,d},
      \widehat{\mathbf{R}}^{\,\mathrm{dir}}_{t,b,d},
      \mathbf{e}_{b},\mathbf{e}_{d},
      \mathbf{m}^{\,\mathrm{active}}_{t,i,b,d}\right),
\end{equation}
where $g$ contains shared encoders, $\mathbf{e}_{b},\mathbf{e}_{d}$ are learned block and variable embeddings, and $\mathbf{m}^{\rm active}$ denotes the remaining scalar residual features. A zero-residual token is appended as candidate $i=0$. Set attention exchanges information among the $K+1$ candidates, followed by a block-variable scoring head:
\begin{equation}
    \boldsymbol{\alpha}_{t,b,d}
    =\operatorname{softmax}
      \left(\operatorname{J5}
      (\{\mathbf{z}_{t,i,b,d}\}_{i=0}^{K})\right).
\end{equation}
The resulting correction is
\begin{equation}
    \widehat{\mathbf{R}}^{\,\mathrm{J5}}_{t,b,d}
    =s\sum_{i=1}^{K}\alpha_{t,i,b,d}\mathbf{R}_{i,b,d};
\end{equation}
weight $\alpha_{t,0,b,d}$ abstains by assigning mass to zero correction and $s$ is a learned global residual scale initialized to one. Candidate order is randomly permuted during training, enforcing set rather than rank semantics.

\subsection{Oracle Imitation and Prediction Loss}

J5 is trained with two complementary objectives. The first is the MSE of the final prediction, which directly constrains whether the combined residual improves the base forecast. The second is the Oracle-imitation loss, which requires the router's candidate weights $\boldsymbol{\alpha}_{t,b,d}$ to approximate the block--variable teacher distribution $\mathbf{q}^*_{t,b,d}$:
\begin{equation}
    \mathcal{L}
    =\operatorname{MSE}(\widehat{\mathbf{Y}}_t,\mathbf{Y}_t)
     +\lambda_{\rm teach}\,
      \operatorname{CE}(q^*_t,\boldsymbol{\alpha}_t),
    \label{eq:loss}
\end{equation}
In the frozen main configuration, the prediction-loss weight is 1; the global teacher-loss scale is 2.0 and the local weight of the block--variable Oracle cross-entropy is 0.2, so the effective coefficient in Eq.~\eqref{eq:loss} is $\lambda_{\rm teach}=2.0\times0.2=0.4$. The prediction loss provides an end-to-end output constraint, while the Oracle cross-entropy provides fine-grained supervision about which historical residual is useful, when, and for which variable. At inference time, J5 uses only the current query, frozen base-model prediction, and train-only memory to produce $\boldsymbol{\alpha}$; it requires neither the true future nor the Oracle teacher.

\subsection{Correction-Strength Selection and Sealed Evaluation}

The learned correction can over-inject residuals. For J5/RATL, we therefore treat $\gamma$ as a scalar forecasting hyperparameter and select it by
\begin{equation}
    \gamma^*
    =\arg\min_{\gamma\in\{0,.1,\ldots,1\}}
       \operatorname{MSE}_{\rm val}
       \left(\widehat{\mathbf{Y}}^{\,0}
       +\gamma\widehat{\mathbf{R}},\mathbf{Y}\right),
    \label{eq:gamma}
\end{equation}
breaking ties toward the smaller $\gamma$. The test set is evaluated only at $\gamma^*$. Equation~\eqref{eq:gamma} applies to J5/RATL, not Direct: Direct is the raw parameter-free similarity-weighted correction fixed at $\gamma=1$ and is not separately $\gamma$-tuned. In the results, RATL denotes J5 with validation-selected $\gamma$. This is ordinary validation-based hyperparameter selection, not predictive-uncertainty calibration or a safety guarantee.

\section{Experiments}

\subsection{Setup}

\paragraph{Datasets and metrics.}
We evaluate 13 public multivariate benchmarks following the provenance convention of iTransformer \cite{liu2024itransformer}: ETTh1, ETTh2, ETTm1, and ETTm2 from the ETT benchmark \cite{zhou2021informer}; ECL, Exchange, Traffic, and Weather as used by Autoformer \cite{wu2021autoformer}; Solar-Energy from LSTNet \cite{lai2018lstnet}; and PEMS03/04/07/08 as evaluated by SCINet \cite{liu2022scinet}. Long-term datasets use horizons $\{96,192,336,720\}$; PEMS uses $\{12,24,36,48\}$, yielding 52 cells. Lookback is 96 and stride is one. ETT follows its official split, the other long-term datasets use chronological 70/10/20 splits, and PEMS uses 60/20/20. Standardization is fit on training data only. We report MSE and MAE, mean and sample standard deviation over seeds $\{1,2,3\}$.

\paragraph{Models and selection.}
The main-experiment base model is iTransformer \cite{liu2024itransformer}. Base-model and J5 checkpoints are selected by minimum validation MSE and always use matched seeds. The frozen RATL configuration uses $K=64$, forecast-horizon blocks of length eight, no similarity feature or similarity prior in J5, and the pruned feature mask described above. Full per-cell configurations, precision tiers, and commands are included in the supplement.

\subsection{Overall Forecasting Accuracy}

Across all 52 cells, RATL reduces MSE by 9.57\% and MAE by 6.21\% on average. Relative to the matched base model, it records 48 wins, 2 ties, and 2 losses; 48 cells improve for every seed, and 47 cells have a positive seed-level 95\% confidence interval. Gains are largest on PEMS (dataset averages of 20.56--24.41\%), but remain positive on ETT, Weather, ECL, Solar, and Traffic. The 2 losses are Exchange-336 ($-2.66\%$) and Exchange-720 ($-8.19\%$).

Table~\ref{tab:main-mse} summarizes the base-model and RATL MSE over every dataset and forecast length in the iTransformer main experiment.
\begin{table*}[t]
\centering
\small
\setlength{\tabcolsep}{3pt}
\textbf{Long-term forecasting}\par
\begin{tabular}{l*{4}{rr}}
\toprule
 & \multicolumn{2}{c}{H=96} & \multicolumn{2}{c}{H=192} & \multicolumn{2}{c}{H=336} & \multicolumn{2}{c}{H=720} \\
\cmidrule(lr){2-3}\cmidrule(lr){4-5}\cmidrule(lr){6-7}\cmidrule(lr){8-9}
Dataset & Base & RATL & Base & RATL & Base & RATL & Base & RATL \\
\midrule
ETTh1 & .387$\pm$.000 & \textbf{.377$\pm$.001} & .439$\pm$.001 & \textbf{.429$\pm$.002} & .480$\pm$.001 & \textbf{.466$\pm$.001} & .491$\pm$.001 & \textbf{.470$\pm$.003} \\
ETTh2 & .304$\pm$.004 & \textbf{.297$\pm$.003} & .380$\pm$.001 & \textbf{.373$\pm$.001} & .421$\pm$.005 & \textbf{.414$\pm$.004} & \textbf{.422$\pm$.001} & \textbf{.422$\pm$.001} \\
ETTm1 & .349$\pm$.001 & \textbf{.327$\pm$.001} & .384$\pm$.001 & \textbf{.362$\pm$.001} & .420$\pm$.001 & \textbf{.399$\pm$.002} & .481$\pm$.002 & \textbf{.463$\pm$.003} \\
ETTm2 & .185$\pm$.000 & \textbf{.178$\pm$.000} & .252$\pm$.001 & \textbf{.247$\pm$.002} & .316$\pm$.002 & \textbf{.309$\pm$.002} & .413$\pm$.001 & \textbf{.409$\pm$.002} \\
Exchange & \textbf{.088$\pm$.001} & \textbf{.088$\pm$.001} & .179$\pm$.000 & \textbf{.178$\pm$.000} & \textbf{.329$\pm$.002} & .338$\pm$.007 & \textbf{.859$\pm$.001} & .929$\pm$.042 \\
Weather & .176$\pm$.002 & \textbf{.165$\pm$.002} & .226$\pm$.002 & \textbf{.216$\pm$.001} & .282$\pm$.001 & \textbf{.274$\pm$.001} & .358$\pm$.001 & \textbf{.353$\pm$.002} \\
ECL & .148$\pm$.001 & \textbf{.134$\pm$.000} & .161$\pm$.001 & \textbf{.150$\pm$.001} & .176$\pm$.003 & \textbf{.163$\pm$.001} & .214$\pm$.004 & \textbf{.192$\pm$.003} \\
Solar & .207$\pm$.002 & \textbf{.200$\pm$.001} & .242$\pm$.002 & \textbf{.228$\pm$.001} & .255$\pm$.002 & \textbf{.234$\pm$.002} & .253$\pm$.000 & \textbf{.232$\pm$.002} \\
Traffic & .400$\pm$.001 & \textbf{.377$\pm$.001} & .418$\pm$.000 & \textbf{.398$\pm$.001} & .433$\pm$.000 & \textbf{.412$\pm$.001} & .466$\pm$.000 & \textbf{.444$\pm$.000} \\
\bottomrule
\end{tabular}
\par\smallskip
\textbf{PEMS traffic forecasting}\par
\begin{tabular}{l*{4}{rr}}
\toprule
 & \multicolumn{2}{c}{H=12} & \multicolumn{2}{c}{H=24} & \multicolumn{2}{c}{H=36} & \multicolumn{2}{c}{H=48} \\
\cmidrule(lr){2-3}\cmidrule(lr){4-5}\cmidrule(lr){6-7}\cmidrule(lr){8-9}
Dataset & Base & RATL & Base & RATL & Base & RATL & Base & RATL \\
\midrule
PEMS03 & .067$\pm$.001 & \textbf{.060$\pm$.000} & .096$\pm$.002 & \textbf{.078$\pm$.000} & .132$\pm$.002 & \textbf{.099$\pm$.001} & .164$\pm$.002 & \textbf{.119$\pm$.001} \\
PEMS04 & .081$\pm$.000 & \textbf{.068$\pm$.000} & .101$\pm$.001 & \textbf{.078$\pm$.000} & .118$\pm$.001 & \textbf{.090$\pm$.001} & .134$\pm$.002 & \textbf{.099$\pm$.002} \\
PEMS07 & .063$\pm$.002 & \textbf{.054$\pm$.002} & .085$\pm$.004 & \textbf{.065$\pm$.002} & .104$\pm$.004 & \textbf{.074$\pm$.002} & .121$\pm$.002 & \textbf{.083$\pm$.001} \\
PEMS08 & .084$\pm$.004 & \textbf{.072$\pm$.001} & .123$\pm$.005 & \textbf{.096$\pm$.001} & .188$\pm$.006 & \textbf{.129$\pm$.003} & .212$\pm$.006 & \textbf{.149$\pm$.005} \\
\bottomrule
\end{tabular}
\caption{Three-seed MSE (mean$\pm$std). RATL denotes the frozen J5 variant with validation-selected $\gamma$. Lower is better.}
\label{tab:main-mse}
\end{table*}

Table~\ref{tab:aggregate-ablation} directly compares Direct and J5 at a fixed correction scale ($\gamma=1$). J5 improves the raw aggregate gain by 1.07 percentage points, supporting the role of learned candidate interaction. The dataset-level breakdown in the supplement further shows that raw Direct is strong on ECL, Traffic, Solar, and PEMS, but degrades results on ETTh2, ETTm2, Exchange, and Weather.

\begin{table}[t]
\centering
\small
\begin{tabular}{lrr}
\toprule
Correction & Mean gain & 95\% CI \\
\midrule
Direct & 7.19\% & [3.84, 10.61] \\
J5 & 8.26\% & [5.18, 11.39] \\
\midrule
J5 $-$ Direct & +1.07 pp & [0.57, 1.62] \\
\bottomrule
\end{tabular}
\caption{Aggregate relative MSE gain over the base at fixed correction scale ($\gamma=1$) across 52 dataset--horizon cells. Confidence intervals are paired cell bootstraps.}
\label{tab:aggregate-ablation}
\end{table}

\paragraph{Correction strength is a consequential hyperparameter.}
For RATL, the selected $\gamma$ is often below one on ETT and Weather, while high-gain Traffic and PEMS cells frequently retain $\gamma=1$. Thus, $\gamma$ is a dataset--horizon--seed-dependent hyperparameter rather than a universal damping constant. Because $\gamma\in[0,1]$, RATL can revert to the original base forecaster when validation rejects the correction.

\subsection{Comparison with Strong Forecasters}

To summarize RATL's transfer behavior across base forecasters, Table~\ref{tab:cross-backbone-summary} collects the iTransformer main panel and the frozen DLinear, PatchTST, TimesNet, and TimeMixer transfer panels. We freeze the best-performing base model and rebuild a train-only residual memory for that exact model. Each row reports the cell-level macro-average gain of RATL relative to its matched frozen base model, rather than comparing absolute forecasting accuracy across backbones. Because the panels differ in coverage and preregistered retrieval keys, this table is intended to show cross-architecture compatibility and its boundaries, not to provide a strict backbone ranking. Complete per-cell absolute metrics and comparisons with independently trained strong forecasters are provided in the supplement.

The retrieval keys used by the different backbones are defined as follows. The iTransformer main experiment uses the per-variable tokens from the final encoder output. The unified DLinear and PatchTST transfer panels in the table use a 12-dimensional input-statistics key consisting of the window's last value, mean, standard deviation, end-to-start difference, and eight-segment average pooling, which keeps the retrieval rule consistent across architectures. PatchTST is also evaluated separately with a native hidden key obtained by temporal pooling of its final encoder patch tokens. TimesNet uses the forecast-horizon hidden states of the final TimesBlock together with the output-head weights to construct per-variable keys. TimeMixer first temporally pools the multiscale hidden states of the final PDM and then averages them across scales. Every key is predefined before testing and uses the same per-variable L2 Top-$K$ retrieval and train-only residual memory. Consequently, differences in the table reflect both the base forecaster and its preregistered key and cannot be attributed to a retrieval representation alone.

The macro-average gain is positive for every backbone in Table~\ref{tab:cross-backbone-summary}, but individual cells can degrade, and the variation in gain shows that transfer depends on the base architecture, dataset, and predefined retrieval key. The TimesNet and TimeMixer hidden-key panels establish compatibility for those evaluated backbone--dataset--key combinations, but they contain no same-backbone input-key arm and therefore do not compare keys or establish hidden-key superiority. PatchTST is additionally evaluated with a native hidden key in a separate matched audit. These results support compatibility under the evaluated settings, not guaranteed improvement. Complete per-cell results, protocol-specific horizons, and integrity audits are provided in the supplement.
\begin{table*}[t]
\centering
\small
\setlength{\tabcolsep}{3pt}
\begin{tabular}{lllrrr}
\toprule
Backbone & Retrieval key & Scope & Cells & MSE gain & MAE gain \\
\midrule
iTransformer & native hidden (encoder) & 13 datasets $\times$ 4 $H$ & 52 & 9.57\% & 6.21\% \\
DLinear & input statistics & 9 datasets $\times$ 1 $H$ & 9 & 5.83\% & 5.75\% \\
PatchTST & input statistics & 9 datasets $\times$ 1 $H$ & 9 & 2.78\% & 1.20\% \\
TimesNet & native hidden (head) & 9 datasets $\times$ 1 $H$ & 9 & 1.42\% & 0.48\% \\
TimeMixer & native hidden (multiscale) & 9 datasets $\times$ 1 $H$ & 9 & 0.58\% & 0.40\% \\
\bottomrule
\end{tabular}
\caption{Cross-backbone summary. Each row reports the macro-average cell gain of RATL over its matched frozen base. The panels use preregistered retrieval keys and different coverage, so the table summarizes transfer scope rather than ranking backbones. DLinear/PatchTST use $H=336$ except Exchange at $H=96$; TimesNet/TimeMixer use the same dataset--horizon scope.}
\label{tab:cross-backbone-summary}
\end{table*}

The supplement also provides source-labeled, published dataset-average comparisons, which show the same broad pattern: RATL is competitive on ETT and Weather, strong on ECL, Solar, Traffic, and PEMS, and weak on Exchange.

\subsection{Robustness, Significance, and Protocol Audit}

Paired test-window bootstrap on representative cells yields positive MSE-gain intervals: [3.95, 4.31]\% for ETTm1-336, [4.89, 5.11]\% for Traffic-336, and [15.03, 16.02]\% for PEMS08-24. In contrast, the intervals for Exchange-336 and Exchange-720 are strictly negative. Analysis of the top $K$ candidates shows that ETTm1 improves as $K$ increases from 16 to 128; Traffic remains stable around $K\in\{16,32,64\}$; and none of the tested $K$ values repairs Exchange-720. The Exchange failures therefore cannot be explained by a single candidate-pool setting.

Unlike traffic, energy, and weather data with stable physical periodicity, we consider Exchange a relatively small financial series that is strongly affected by exogenous shocks and exhibits clear regime changes. Its exchange-rate levels, volatility, and cross-variable relationships may change over time, while stable daily or weekly periodicity is relatively weak. Thus, even when two input windows are similar in historical shape or backbone representation, their subsequent exchange-rate changes and base-model errors need not be similar. The RATL assumption that similar contexts have transferable residual patterns is more likely to fail on this dataset, causing historical residuals retrieved at test time to be mismatched in direction or magnitude. Long horizons also reduce the number of effective windows available for validation selection, further increasing uncertainty in correction-strength estimation.

Because retrieved residuals may not provide reliable corrections in settings with weak periodicity or strong distribution shift, RATL includes two predefined fallback-to-base mechanisms. Locally, J5 adds an explicit zero-residual candidate at every block--variable position, allowing the router to reduce or cancel the residual correction at that position. Globally, the validation candidate set includes $\gamma=0$, so the final output can revert completely to the frozen base forecaster when validation evidence does not support a nonzero correction. These mechanisms reduce the risk of negative transfer but do not provide a non-degradation or safety guarantee.

\section{Conclusion and Future Work}

We propose RATL, which causally retrieves historical residuals from a frozen base forecaster and uses a block--variable router to select and combine candidate residuals. The iTransformer main experiment shows that model-specific historical errors can serve as reusable inference-time memory; the DLinear, PatchTST, TimesNet, and TimeMixer transfer panels further provide evidence that the interface is compatible with multiple architectures in the evaluated settings.

RATL's gains vary with the base forecaster, dataset, and retrieval-key definition. For datasets with limited data, weak periodicity, and more frequent changes driven by exogenous factors, historical residuals from similar contexts need not transfer to the test stage. The zero-residual candidate and $\gamma=0$ fallback can mitigate negative transfer but provide no non-degradation or safety guarantee. At the same time, exact retrieval and long multivariate residual memories incur growing computational and storage costs as the numbers of training windows and variables and the forecast length increase. Semantic representations of exogenous factors could be incorporated into retrieval-key vectors.

Future work will focus on uncertainty-aware abstention based on candidate disagreement and distribution drift, sample-adaptive correction strength, and validation-safe retrieval-key selection. It will also explore approximate nearest-neighbor search, residual quantization, prototype memories, and dynamic memory compression to reduce deployment costs. Further studies should test the cross-regime stability of residual patterns across broader backbones and real operating environments, and assess applicability to transportation, energy, and industrial forecasting together with domain constraints, online monitoring, and conservative fallback strategies.

\bibliography{references}

\end{document}